\pdfoutput=1
\documentclass{article}

\usepackage[numbers]{natbib}

\usepackage[final]{neurips_2021_ml4ps}

\usepackage[utf8]{inputenc} 
\usepackage[T1]{fontenc}    
\usepackage{hyperref}       
\usepackage{url}            
\usepackage{booktabs}       
\usepackage{amsfonts}       
\usepackage{nicefrac}       
\usepackage{microtype}      
\usepackage{xcolor}         
\usepackage{bm}
\usepackage{authblk}

\usepackage[pdftex]{graphicx}
\usepackage[toc,page]{appendix}
\usepackage{amsmath}
\usepackage[textsize=scriptsize]{todonotes}

\title{S3RP: Self-Supervised Super-Resolution and Prediction for Advection-Diffusion Process}

\author{\textbf{
 Chulin Wang; Kyongmin Yeo; Xiao Jin; Andres Codas; Levente J. Klein; Bruce Elmegreen} \\
 IBM T.J. Watson Research Center \\
 Yorktown Heights, New York 10598 \\
 wangc@ibm.com; kyeo@us.ibm.com; Xiao.Jin@ibm.com; andrescodas@br.ibm.com; \{kleinl, bge\}@us.ibm.com \\
}

\begin{document}

\maketitle

\begin{abstract}
We present a super-resolution model for an advection-diffusion process with limited information. While most of the super-resolution models assume high-resolution (HR) ground-truth data in the training, in many cases such HR dataset is not readily accessible. Here, we show that a Recurrent Convolutional Network trained with physics-based regularizations is able to reconstruct the HR information without having the HR ground-truth data. Moreover, considering the ill-posed nature of a super-resolution problem, we employ the Recurrent Wasserstein Autoencoder to model the uncertainty. 
\end{abstract}

\section{Introduction}
\label{sec:intro}

Super-resolution (SR) reconstruction of an advection-diffusion process has a direct relevance to many important applications in atmospheric and environmental problems. 
SR and prediction are two of the most important aspects of down scaling climate/weather modelling and satellite observations, where ground truth is not readily available. 
Generating SR from low resolution data is an ill-posed problem as multiple high resolution solutions may exist corresponding to a low resolution data. 

SR has been studied in the machine learning community with various methods. Deterministic methods include SRCNN\cite{dong2015image} and more recent development ESRGAN\cite{wang2018esrgan}, among others. However such methods are inherently deterministic, which predict the \textit{mean} of all possible HR and tend to lose fine random features. Probabilistic models such as SRFlow\cite{lugmayr2020srflow} based on Normalizing Flow\cite{rezende2015variational} and PULSE\cite{menon2020pulse} based on a pretrained StyleGAN\cite{karras2019style} have been successful in generating a distribution of HR, but are difficult to generalize to an arbitrary dataset.

Probabilistic time series predictions have been studied in a number of architectures, for example, MoCoGAN\cite{tulyakov2018mocogan}, S3VAE\cite{zhu2020s3vae}, and VideoFlow\cite{kumar2019videoflow}. Most of the prior works have pleasing results with toy examples, but perform poorly in complex scenarios due to a lack of domain knowledge.

Multiple physics-informed neural networks have been proposed for SR\cite{vandal2017deepsd,wang2020physics,gao2021super}, time series generation\cite{guen2020disentangling,sonderby2020metnet,wang2020towards} and spatial-temporal super-resolution\cite{esmaeilzadeh2020meshfreeflownet,xie2018tempogan}. To the best of our knowledge, none of the prior works have combined probabilistic SR and time series prediction in physics-informed neural networks.

We present our work of self-supervised super-resolution and prediction (S3RP) neural networks that address all of the above mentioned issues in one physics-informed neural network architecture. Our method has the following advantages:
\begin{itemize}
    \itemsep -0.2 em 
    \item Address the common problem of the lack of ground-truth HR data.
    \item Model the uncertainty with a probabilistic model.
    \item Achieve spatial SR and temporal prediction in the same neural network architecture that is constrained by physics equation.
\end{itemize}


\section{Method}

\subsection{Problem setup}
\label{sec:sims}

Here, we consider the following 2-dimensional advection-diffusion equation;
\begin{equation}
    \partial_t c + \nabla \cdot c \bm{u} = \nabla \cdot (\bm{K} \cdot \nabla c) + Q,
    \label{eq:ade}
\end{equation}

where $c$ is the concentration, $\bm{u}$ is the velocity field, $\bm{K}$ is the eddy-diffusivity tensor, and $Q$ is the source term. The wind field is generated by a multiscale Langevin process as described in \cite{Yeo19}. The wind velocity satisfies the following mass conservation equation;
\begin{equation}
    \nabla \cdot \bm{u} = 0.
    \label{eq:div_free}
\end{equation}
In many real-life problems, $\bm{u}$ and $c$ can be obtained from satellite images or weather models. However, $Q$ and $\bm{K}$ are generally unknown. Hence, we also assume that we only have the low-resolution (LR) data for $c$ and $\bm{u}$, not $Q$ and $\bm{K}$.

We assume the LR data is defined in a rectangular mesh with a uniform spacing; $\mathcal{W}^L = \bm{s}^L_1 \bigotimes \bm{s}^L_2$, where $\bm{s}^L_i = \{s^0_i+j \delta s_L,~j=0,\cdots,N_L -1 \}$. Let $\bm{x}_t \in \mathbb{R}^{N_L\times N_L \times 3}$ be the LR data used for training, $\bm{y}_t\in \mathbb{R}^{N_H\times N_H \times 3}$ be the HR ground-truth, and $\hat{\bm{y}}_t\in \mathbb{R}^{N_H\times N_H \times 3}$ be the SR solution by the neural network. Note that $\bm{x}_t$, $\bm{y}_t$ and $\hat{\bm{y}}_t$ consist of three channels; two for the velocity components and one for the concentration. We assume that the SR image downsamples correctly with respect to the data, \emph{i.e.}, $\textrm{DS}(\hat{y}_t) \approx x_t$ \cite{menon2020pulse}, where $\textrm{DS}(\cdot)$ is the downsampling process of natural images. If we assume a simple spatial average, the high-resolution grid system, $\mathcal{W}^H$, is naturally defined by $\textrm{DS}$.





\subsection{Deep Learning Model}

Consider the following stochastic advection equation,
\begin{equation}
    \partial_t c = - \nabla \cdot c \bm{u} + {\epsilon}.
    \label{eq:spe}
\end{equation}
Here, $\epsilon$ is a random variable due to the uncertainty in $Q$ and $\bm{K}$. We impose regularizations for the deterministic terms and employ the Recurrent Wasserstein Autoencoder (WAE) to model the stochasticity.

\paragraph{Physics regularization}
For the physics regularization, we consider the deterministic terms of Eq.~(\ref{eq:spe}) as well as the mass conservation of the ambient fluid;
\begin{equation}
    \mathcal{L}_\textrm{phy} = \| \partial_t c + \nabla \cdot c \bm{u} \|_2^2 + 
    \beta \cdot \| \nabla \cdot \bm{u} \|_2^2.
    \label{eq:phys_loss}
\end{equation}

\paragraph{Recurrent WAE}
Here, we consider the following data generating distribution,
\begin{equation}
    p(\bm{x}_{0:t}, \bm{z}_{0:t}) = \prod_{i=0}^t{p_\theta(\bm{x}_{i} | \bm{z}_i) p(\bm{z}_i | \bm{z}_{0:i-1})}, \label{eq:generative}
\end{equation}
in which $\bm{z}_t$ is a latent variable. Then, as shown in \cite{han2021disentangled}, minimizing the following loss function corresponds to minimizing a Wasserstein distance between the data generating distribution and the distribution of a deep generative model,
\begin{equation}
    \mathcal{L}_\mathrm{WAE} = \sum_{i=1}^{t} {
       \mathbb{E}_{q(\bm{z}_i | \bm{z}_{0:i-1}, \bm{x}_{i-1})}[l(\bm{x}_{i}, G(\bm{z}_i))] 
       + \lambda \sum_{i=0}^t \mathcal{D}\left(q_{\bm{z}_i | \bm{z}_{0:i-1}}, p_{\bm{z}_i | \bm{z}_{0:i-1}}\right)
    },
    \label{eq:D_WAE}
\end{equation}
in which $l(\cdot,\cdot)$ 
denotes a cost function, $\mathcal{D}(q,p)$ is a divergence between distributions, and $p_{\bm{z}_i | \bm{z}_{0:i-1}}$ is a prior distribution.
Here, the posterior distribution, $q(\bm{z}_i | \bm{z}_{0:i-1}, \bm{x}_i)$, and the generator, $G(\bm{z}_i)$, are parameterized by neural networks. The $l^2$-norm is used for the cost function, $l(\bm{x}_{t+1},G(\bm{z}_t)) = \| \bm{x}_{t+1} - G(\bm{z}_t)\|^2_2$. $G(\bm{z}_t)$ consists of a super-resolution decoder followed by the downsampling,
\[
G(\bm{z}_t) = \textrm{DS}(\hat{\bm{y}}_{t}),
\]
in which $\hat{\bm{y}}_{t} = \Psi_\mathrm{dec}(\bm{z}_t)$ and $\Psi_\mathrm{dec}$ denotes a decoder neural network. The maximum mean discrepancy\cite{dziugaite2015training} is used for the divergence, $\mathcal{D}$.

WAE has a similar structure with the variational autoencoder (VAE). However, unlike VAE, WAE does not require explicitly defining the emission probability, $p(\bm{x}_{t}|\bm{z}_t)$, which is advantageous when $\bm{x}_{t}$ has a complex correlation structure, such as the physical constraints in the current problem setup. Finally, the loss function is given as the sum of WAE and the physics loss;
\begin{equation}
    \mathcal{L} = \mathcal{L}_\mathrm{WAE}+\gamma\mathcal{L}_\mathrm{phy}.
\end{equation}

\begin{figure}[htbp]
    \centering
    \includegraphics[width=0.6\linewidth]{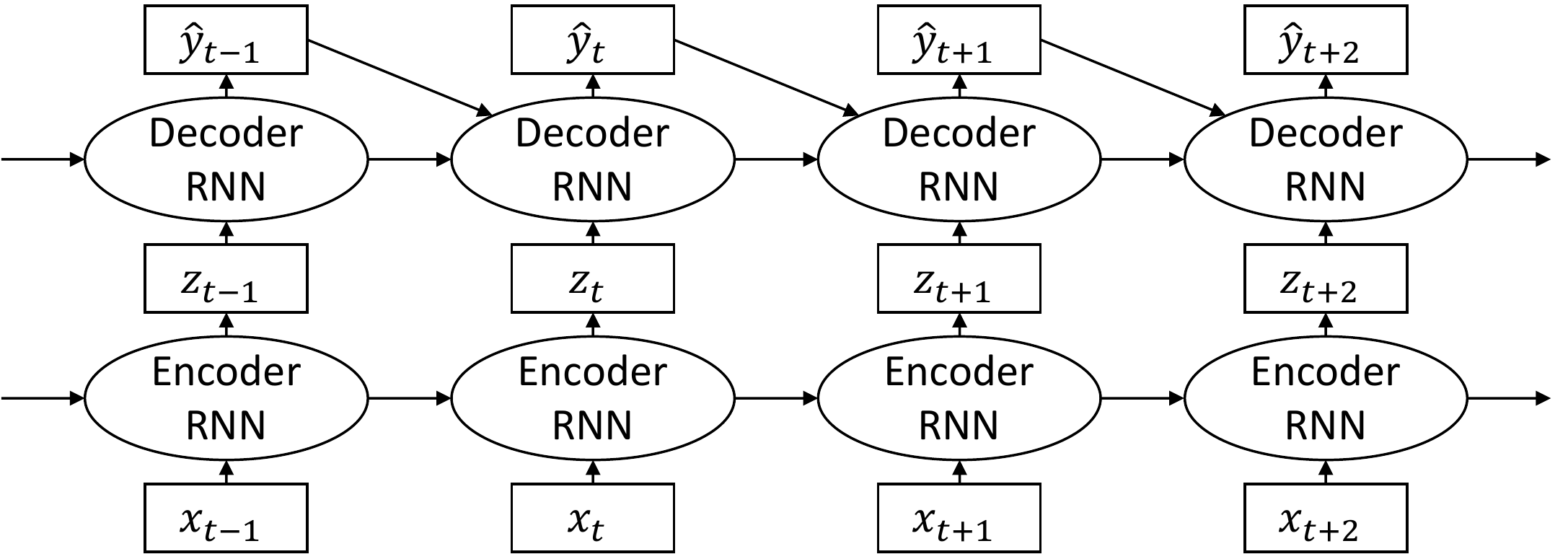}
    \caption{
        A sketch of our Recurrent WAE neural network. The recurrent encoder is probabilistic, representing the conditional distribution of $\bm{z}_t$ given the past LR data $\bm{x}_{0:t}$. The recurrent decoder is deterministic, and is a function of $\bm{z}_{0:t}$ and the last HR output $\hat{y}_{t-1}$, and generates the new output $\hat{y}_{t}$.
    }
    \label{fig:network}
\end{figure}

\paragraph{Neural network architecture.} 
\label{sec:arch}
A diagram of our neural network is shown in Fig.~\ref{fig:network}. The conditional probabilities are approximated by RNNs in both the encoding and decoding process. The initial condition, $\hat{y}_0$, is generated by a bilinear upsampling of the LR data. We use a LSTM for the variational encoder, and we use a PhyCell (see Section~\ref{sec:phycell}) in conjunction with a LSTM for the decoder.

\section{Experiments}

\subsection{Dataset}

Following Section~\ref{sec:sims}, we generate 10 simulations with different $\bm u$ and $\bm{K}$. Within each simulation, 16 concentration fields are computed for 16 source locations, separately. The final concentration data is generated by a random combination of the concentration fields due to the linearity. The HR simulations is performed for the mesh size $N_H = 128$. The LR training data set $\bm{x}_t$ is subsequently generated by a spatial average over $8\times 8$ pixels in HR, resulting in a input data $\bm{x}_t \in \mathbb{R}^{16\times 16 \times 3}$.

\subsection{Baseline and Models}
\label{sec:baseline}

{\bf Bicubic.} 
We use bicubic upsampling to generate SR images as the baseline.\\
{\bf S3RP model (Interpolation model).} We use $x_{0:t}$ for SR reconstruction of $y_{0:t}$ as shown in Eq.~(\ref{eq:generative})\\
{\bf S3RP model (Other variations).} The same model can be reconfigured to perform SR of $\bm{u}$ as well as SR and forecast of $c$ (\textit{c}-only model). Alternatively, the same model can perform SR and forecast of $(c,\bm{u})$ (full extrapolation model). Detailed discussion can be found in Appendix~\ref{sec:suppl_model}



\subsection{Metric}

We evaluate the model by a Monte Carlo (MC) simulation with 100 samples.
Then, the expectation and the prediction intervals are estimated. The mean squared error (MSE) is evaluated between the expectation and the HR ground truth. The empirical coverage probability is also shown to provide a quantitative assessment of the probabilistic model.
Finally, we also evaluate the physics error of the model output and verify that the predictions generated by the model is indeed physically consistent.

\subsection{Results}
\begin{figure}[htbp]
    \centering
    \includegraphics[width=0.7\linewidth]{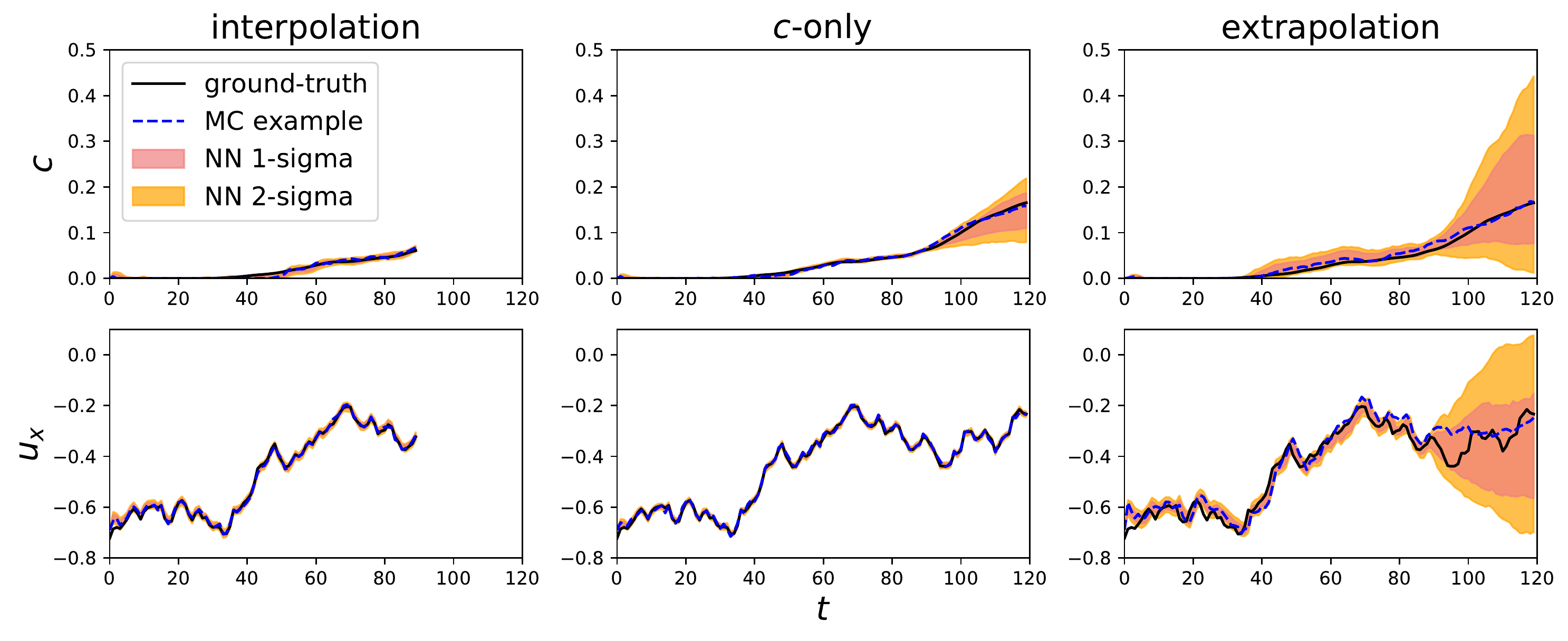}
    \caption{
        Comparison between the ground-truth and the model output at an arbitrary coordinate. A sample from the MC simulation is shown as the dashed line. Each column represent one of the variations of the model. 
    }
    \label{fig:2_sigma}
\end{figure}

We evaluate the model performance by using a hold-out LR data for 90 time steps, $\bm{x}\in\mathbb{R}^{90\times16\times16\times3}$, and comparing the output with the ground-truth HR. The results of the interpolation model are summarized in Table~\ref{table:compare_interp}. 
It is shown that our model outperforms the baseline. Since our model is a generative model that computes the probability distribution, we also showed the empirical coverage probability (ECP)
for a quantitative comparison in Table~\ref{table:compare_interp}. More discussions about the estimated probability distribution is provided in  Appendix~\ref{sec:error_std}. In Fig.~\ref{fig:interp_vis}, the model output of concentration $c$ is visualized, together with the estimated standard deviation. 
The model also outperforms the baseline in capturing the physics process. Fig.~\ref{fig:phys_error} shows a snapshot of the physics errors. In particular, it is shown that the SR solution of our model well satisfies the mass conservation condition
shown in Eq.~(\ref{eq:div_free}),
compared to the baseline. In Table~\ref{table:compare_interp}, it is shown that $\epsilon_{div}$ of our model is an order of magnitude smaller than the baseline.
As mentioned in Section~\ref{sec:baseline}, the models can be set up to accommodate for different input settings. We show the range of outputs for an a given coordinate generated by all 3 variations of the probabilistic model in Fig.~\ref{fig:2_sigma}.
In Fig~\ref{fig:2_sigma}, for $c$-only and extrapolation models, the LR data is provided only up to $t=90$, and the model makes a probabilistic forecast by a Monte Carlo simulation for the next 30 steps. It is shown that the prediction interval of the extrapolation mode becomes much larger than that of the $c$-only mode due to the uncertainties in the future wind condition.

\begin{table}[htbp]
  \caption{Comparison of our S3RP model (interpolation mode) and baseline. The $1\sigma$ and $2\sigma$ coverage denote 68\% and 95\% prediction intervals. The physics errors, $\epsilon_\mathrm{adv-diff}$ and  $\epsilon_\mathrm{div}$, are defined in Appendix~\ref{sec:phys_error}. The best results are in boldface.}
  \label{table:compare_interp}
  \centering
  \scalebox{0.8}{
  \begin{tabular}{cccccc}
    \toprule
     & MSE & $1\sigma$ coverage & $2\sigma$ coverage & $\epsilon_\mathrm{adv-diff}$ & $\epsilon_\mathrm{div}$\\
    \midrule
    S3RP & \bf 1.38E-4 & \bf 65.5\% & \bf 82.2\% & \bf 1.69e-6 & \bf 6.28E-6\\
    Bicubic & 3.97E-4 & - & - & 1.83E-6 & 6.20E-5\\
    \bottomrule
  \end{tabular}
  }
\end{table}

\begin{figure}[htbp]
    \centering
    \begin{minipage}{0.45\textwidth}
        \centering
        \includegraphics[width=0.8\textwidth]{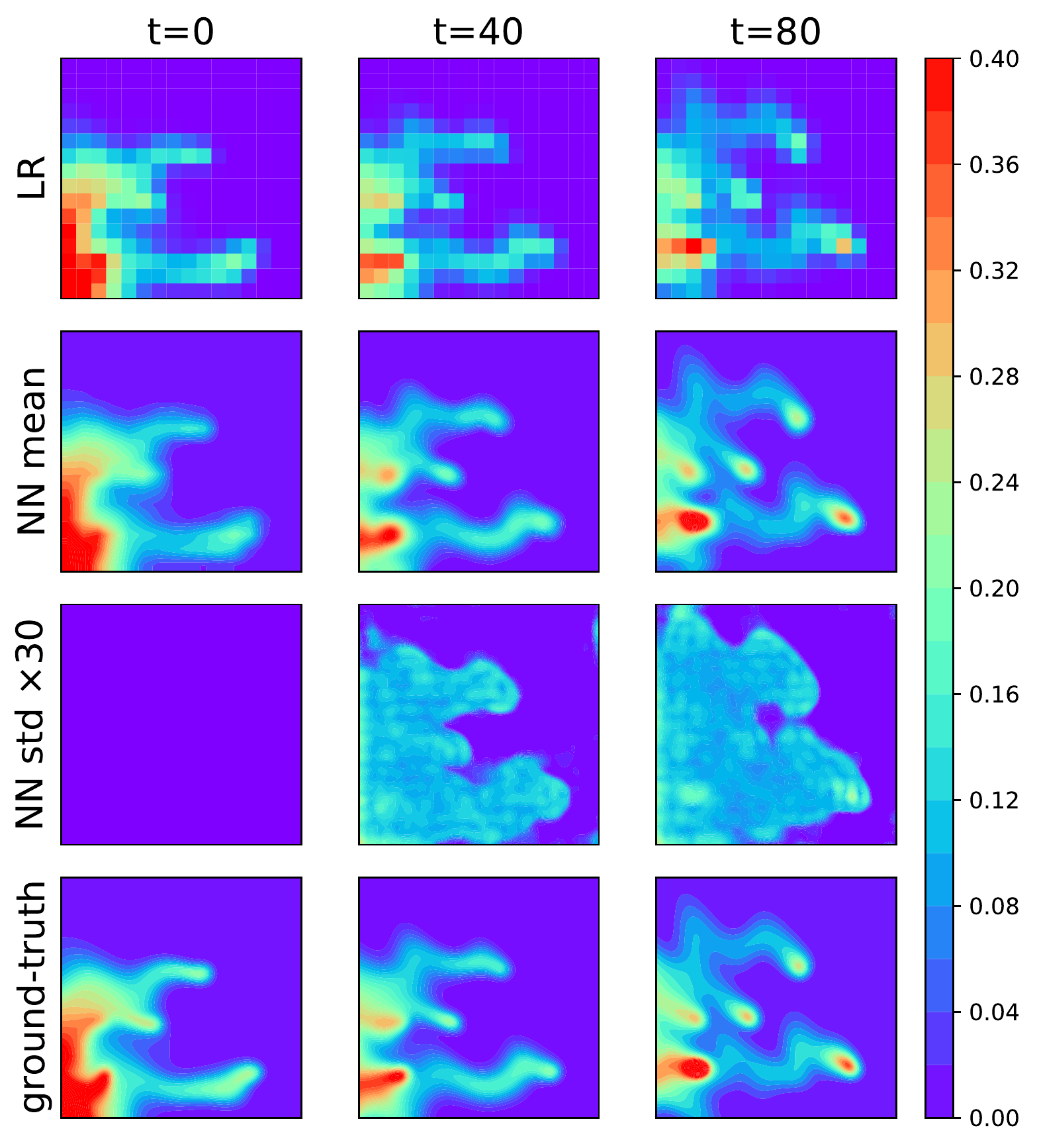} 
        \caption{Visualization of the S3RP model at different timesteps. The model (2nd row) is able to perform $8\times$ SR accurately as well as capture the uncertainty (3rd row).}
        \label{fig:interp_vis}
    \end{minipage}\hfill
    \begin{minipage}{0.50\textwidth}
        \centering
        \includegraphics[width=0.95\textwidth]{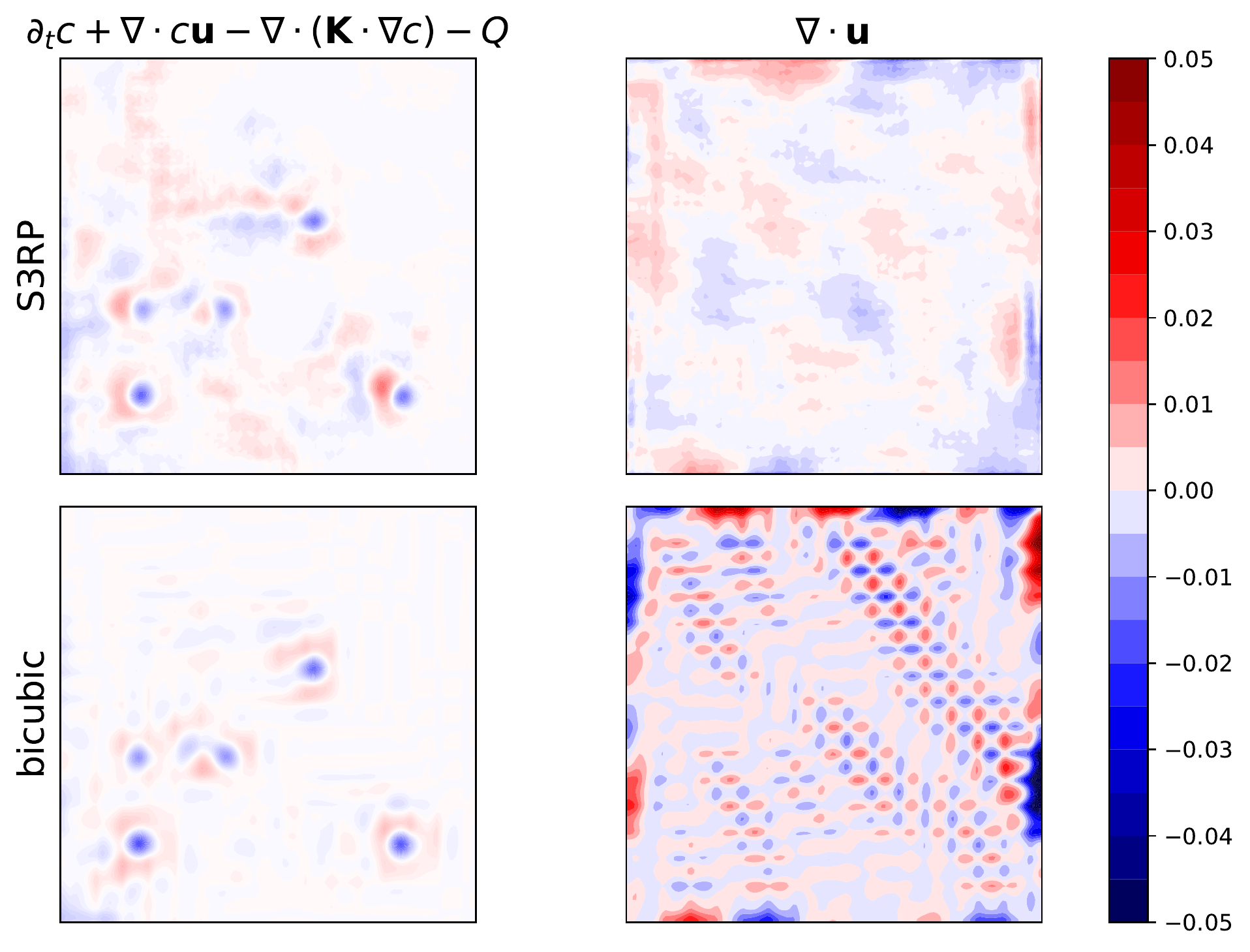}
        \caption{
        A snapshot of the spatial distribution of the physics errors. The first and second rows correspond to S3RP and the bicubic interpolations, respectively. 
        }
        \label{fig:phys_error}
    \end{minipage}
\end{figure}

\section{Conclusions}
We propose a deep learning model, S3RP, which achieves both super-resolution and prediction for advection-diffusion process. The training is self-supervised assuming no access to HR ground-truth. The uncertainty is estimated by using WAE. The model also embeds physics equations by using a physics regularization and a specially designed module, PhyCell, in the network. As the result shows, the model performs better in all metrics against the baseline. We expect the same framework can be applied to many real-world physics problems, especially where only LR data is available.


\section*{Broader Impact} 
We propose to approach the SR problem from a probabilistic formulation, because in a SR problem uniqueness of the solution is not guaranteed. Moreover, the model enables dense spatial-temporal predictions when high resolution ground truth doesn't exist but the governing physics are well understood. Such problems are commonly encountered in Earth and Environmental Sciences, such as pollution prediction and source identification from satellite images or for autonomous vehicle/robot navigation  under hazardous conditions.

\bibliographystyle{unsrt}
\bibliography{refs}

\begin{thebibliography}{10}

\bibitem{dong2015image}
Chao Dong, Chen~Change Loy, Kaiming He, and Xiaoou Tang.
\newblock Image super-resolution using deep convolutional networks.
\newblock {\em IEEE transactions on pattern analysis and machine intelligence},
  38(2):295--307, 2015.

\bibitem{wang2018esrgan}
Xintao Wang, Ke~Yu, Shixiang Wu, Jinjin Gu, Yihao Liu, Chao Dong, Yu~Qiao, and
  Chen Change~Loy.
\newblock Esrgan: Enhanced super-resolution generative adversarial networks.
\newblock In {\em Proceedings of the European conference on computer vision
  (ECCV) workshops}, pages 0--0, 2018.

\bibitem{lugmayr2020srflow}
Andreas Lugmayr, Martin Danelljan, Luc Van~Gool, and Radu Timofte.
\newblock Srflow: Learning the super-resolution space with normalizing flow.
\newblock In {\em European Conference on Computer Vision}, pages 715--732.
  Springer, 2020.

\bibitem{rezende2015variational}
Danilo Rezende and Shakir Mohamed.
\newblock Variational inference with normalizing flows.
\newblock In {\em International conference on machine learning}, pages
  1530--1538. PMLR, 2015.

\bibitem{menon2020pulse}
Sachit Menon, Alexandru Damian, Shijia Hu, Nikhil Ravi, and Cynthia Rudin.
\newblock Pulse: Self-supervised photo upsampling via latent space exploration
  of generative models.
\newblock In {\em Proceedings of the ieee/cvf conference on computer vision and
  pattern recognition}, pages 2437--2445, 2020.

\bibitem{karras2019style}
Tero Karras, Samuli Laine, and Timo Aila.
\newblock A style-based generator architecture for generative adversarial
  networks.
\newblock In {\em Proceedings of the IEEE/CVF Conference on Computer Vision and
  Pattern Recognition}, pages 4401--4410, 2019.

\bibitem{tulyakov2018mocogan}
Sergey Tulyakov, Ming-Yu Liu, Xiaodong Yang, and Jan Kautz.
\newblock Mocogan: Decomposing motion and content for video generation.
\newblock In {\em Proceedings of the IEEE conference on computer vision and
  pattern recognition}, pages 1526--1535, 2018.

\bibitem{zhu2020s3vae}
Yizhe Zhu, Martin~Renqiang Min, Asim Kadav, and Hans~Peter Graf.
\newblock S3vae: Self-supervised sequential vae for representation
  disentanglement and data generation.
\newblock In {\em Proceedings of the IEEE/CVF Conference on Computer Vision and
  Pattern Recognition}, pages 6538--6547, 2020.

\bibitem{kumar2019videoflow}
Manoj Kumar, Mohammad Babaeizadeh, Dumitru Erhan, Chelsea Finn, Sergey Levine,
  Laurent Dinh, and Durk Kingma.
\newblock Videoflow: A conditional flow-based model for stochastic video
  generation.
\newblock {\em arXiv preprint arXiv:1903.01434}, 2019.

\bibitem{vandal2017deepsd}
Thomas Vandal, Evan Kodra, Sangram Ganguly, Andrew Michaelis, Ramakrishna
  Nemani, and Auroop~R Ganguly.
\newblock Deepsd: Generating high resolution climate change projections through
  single image super-resolution.
\newblock In {\em Proceedings of the 23rd acm sigkdd international conference
  on knowledge discovery and data mining}, pages 1663--1672, 2017.

\bibitem{wang2020physics}
Chulin Wang, Eloisa Bentivegna, Wang Zhou, Levente Klein, and Bruce Elmegreen.
\newblock Physics-informed neural network super resolution for
  advection-diffusion models.
\newblock {\em arXiv preprint arXiv:2011.02519}, 2020.

\bibitem{gao2021super}
Han Gao, Luning Sun, and Jian-Xun Wang.
\newblock Super-resolution and denoising of fluid flow using physics-informed
  convolutional neural networks without high-resolution labels.
\newblock {\em Physics of Fluids}, 33(7):073603, 2021.

\bibitem{guen2020disentangling}
Vincent~Le Guen and Nicolas Thome.
\newblock Disentangling physical dynamics from unknown factors for unsupervised
  video prediction.
\newblock In {\em Proceedings of the IEEE/CVF Conference on Computer Vision and
  Pattern Recognition}, pages 11474--11484, 2020.

\bibitem{sonderby2020metnet}
Casper~Kaae S{\o}nderby, Lasse Espeholt, Jonathan Heek, Mostafa Dehghani,
  Avital Oliver, Tim Salimans, Shreya Agrawal, Jason Hickey, and Nal
  Kalchbrenner.
\newblock Metnet: A neural weather model for precipitation forecasting.
\newblock {\em arXiv preprint arXiv:2003.12140}, 2020.

\bibitem{wang2020towards}
Rui Wang, Karthik Kashinath, Mustafa Mustafa, Adrian Albert, and Rose Yu.
\newblock Towards physics-informed deep learning for turbulent flow prediction.
\newblock In {\em Proceedings of the 26th ACM SIGKDD International Conference
  on Knowledge Discovery \& Data Mining}, pages 1457--1466, 2020.

\bibitem{esmaeilzadeh2020meshfreeflownet}
Soheil Esmaeilzadeh, Kamyar Azizzadenesheli, Karthik Kashinath, Mustafa
  Mustafa, Hamdi~A Tchelepi, Philip Marcus, Mr~Prabhat, Anima Anandkumar,
  et~al.
\newblock Meshfreeflownet: a physics-constrained deep continuous space-time
  super-resolution framework.
\newblock In {\em SC20: International Conference for High Performance
  Computing, Networking, Storage and Analysis}, pages 1--15. IEEE, 2020.

\bibitem{xie2018tempogan}
You Xie, Erik Franz, Mengyu Chu, and Nils Thuerey.
\newblock tempogan: A temporally coherent, volumetric gan for super-resolution
  fluid flow.
\newblock {\em ACM Transactions on Graphics (TOG)}, 37(4):1--15, 2018.

\bibitem{Yeo19}
Kyongmin Yeo, Youngdeok Hwang, Xiao Liu, and Jayant Kalagnanama.
\newblock Development of $hp$-inverse model by using generalized polynomial
  chaos.
\newblock {\em Computer Methods in Applied Mechanics and Engineering},
  347:1--20, 2019.

\bibitem{han2021disentangled}
Jun Han, Martin~Renqiang Min, Ligong Han, Li~Erran Li, and Xuan Zhang.
\newblock Disentangled recurrent wasserstein autoencoder.
\newblock {\em arXiv preprint arXiv:2101.07496}, 2021.

\bibitem{dziugaite2015training}
Gintare~Karolina Dziugaite, Daniel~M Roy, and Zoubin Ghahramani.
\newblock Training generative neural networks via maximum mean discrepancy
  optimization.
\newblock {\em arXiv preprint arXiv:1505.03906}, 2015.

\end{thebibliography}

\clearpage

\begin{appendices}
\appendix

\section{PhyCell}
\label{sec:phycell}

As studied in \cite{guen2020disentangling}, recurrent neural networks (RNN) is not efficient in video prediction. PhyCell is proposed as a special module that encodes an arbitrary partial differential equation (PDE). We implement the same structure in the decoder $\Psi_\mathrm{dec}$ and observed better extrapolation results of Eq.~\ref{eq:ade}.

\section{Architecture Details}
\label{sec:suppl_arch}
\begin{figure}[htbp]
    \centering
    \begin{minipage}{0.30\textwidth}
        \centering
        \includegraphics[width=1.0\textwidth]{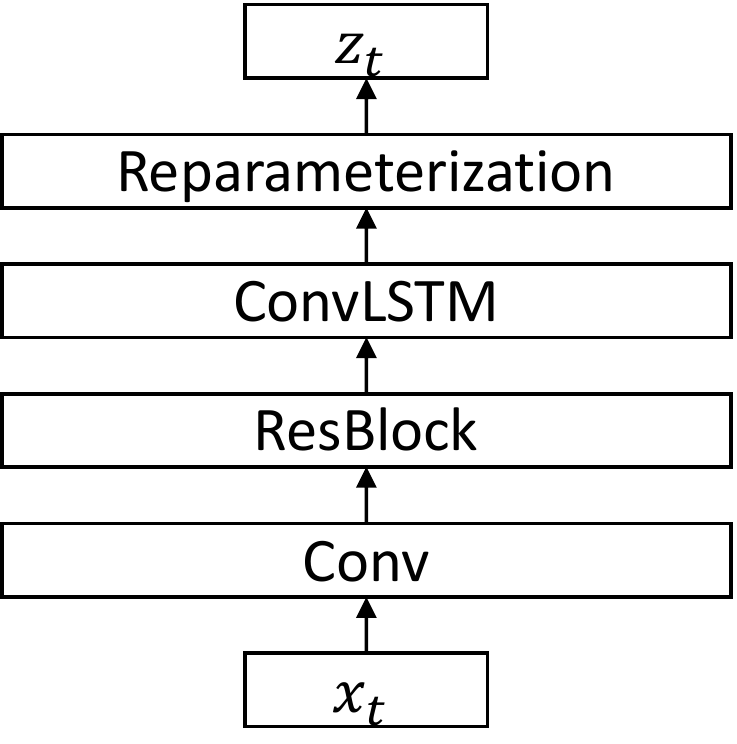} 
        \caption{The variational encoder is a probabilistic model. It takes as input the LR data, $x_t$, and outputs a sample of the latent variable $z_t$, which is done by reparameterization a unit gaussian with the mean and standard deviation predicted by the model. The ConvLSTM handles the memory of previous LR data $x_{0:t-1}$.}
        \label{fig:encoder}
    \end{minipage}\hfill
    \begin{minipage}{0.60\textwidth}
        \centering
        \includegraphics[width=1.0\textwidth]{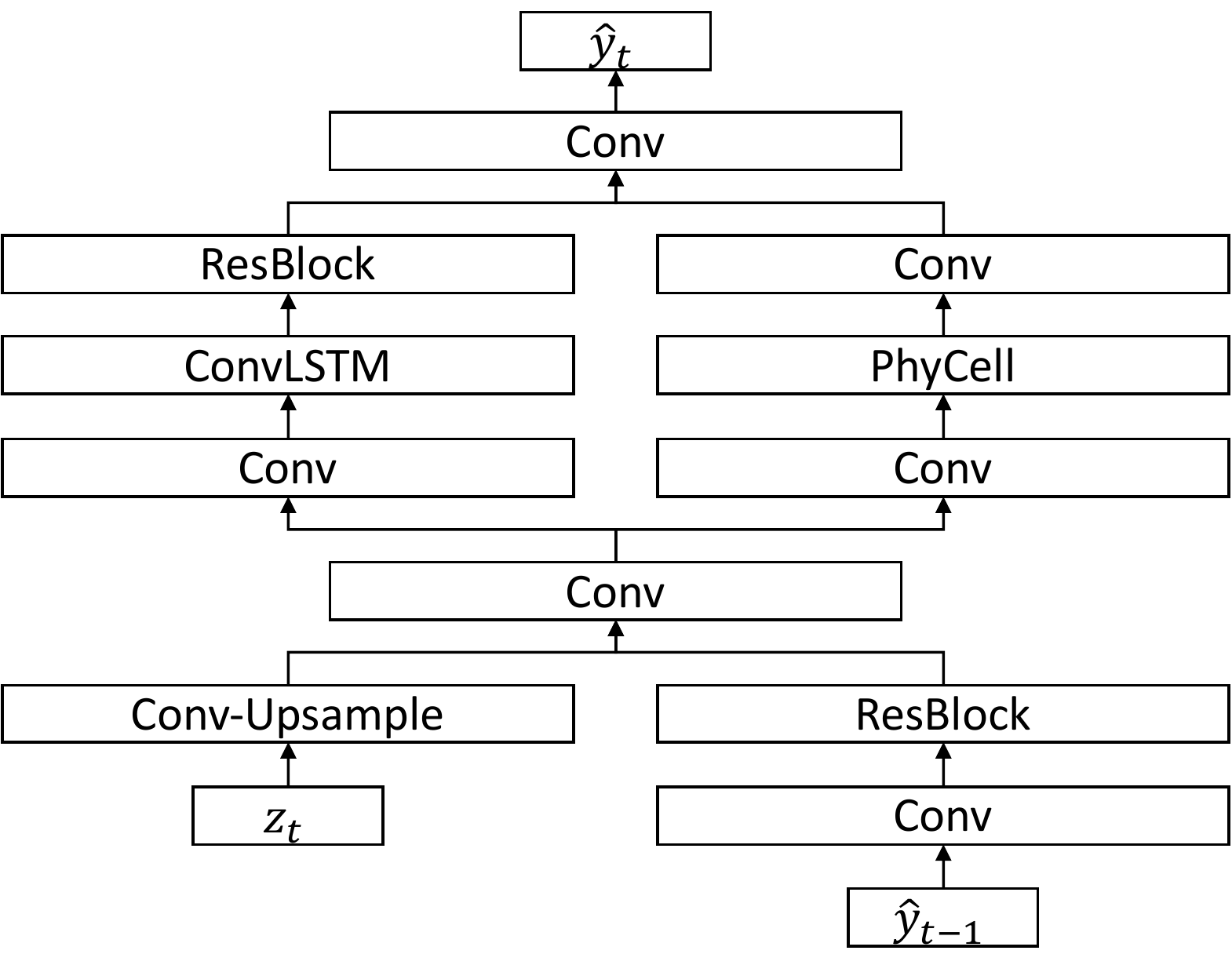}
        \caption{The variational decoder is deterministic (Dirac delta probability distribution). The input of the decoder is the latent variable $z_t$ and the previous HR output $\hat{y}_{t-1}$, and outputs the new HR output $\hat{y}_t$. The decoder is comprised of two branches, the PhyCell that captures the PDE, and the ConvLSTM that takes care of the residual dynamics.}
        \label{fig:decoder}
    \end{minipage}
\end{figure}

The detailed encoder and decoder structure is illustrated in Figs.~\ref{fig:encoder} and \ref{fig:decoder}.

\section{Other Model Variations}
\label{sec:suppl_model}
\subsection{Full extrapolation model}
\begin{figure}[htbp]
    \centering
    \includegraphics[width=0.7\linewidth]{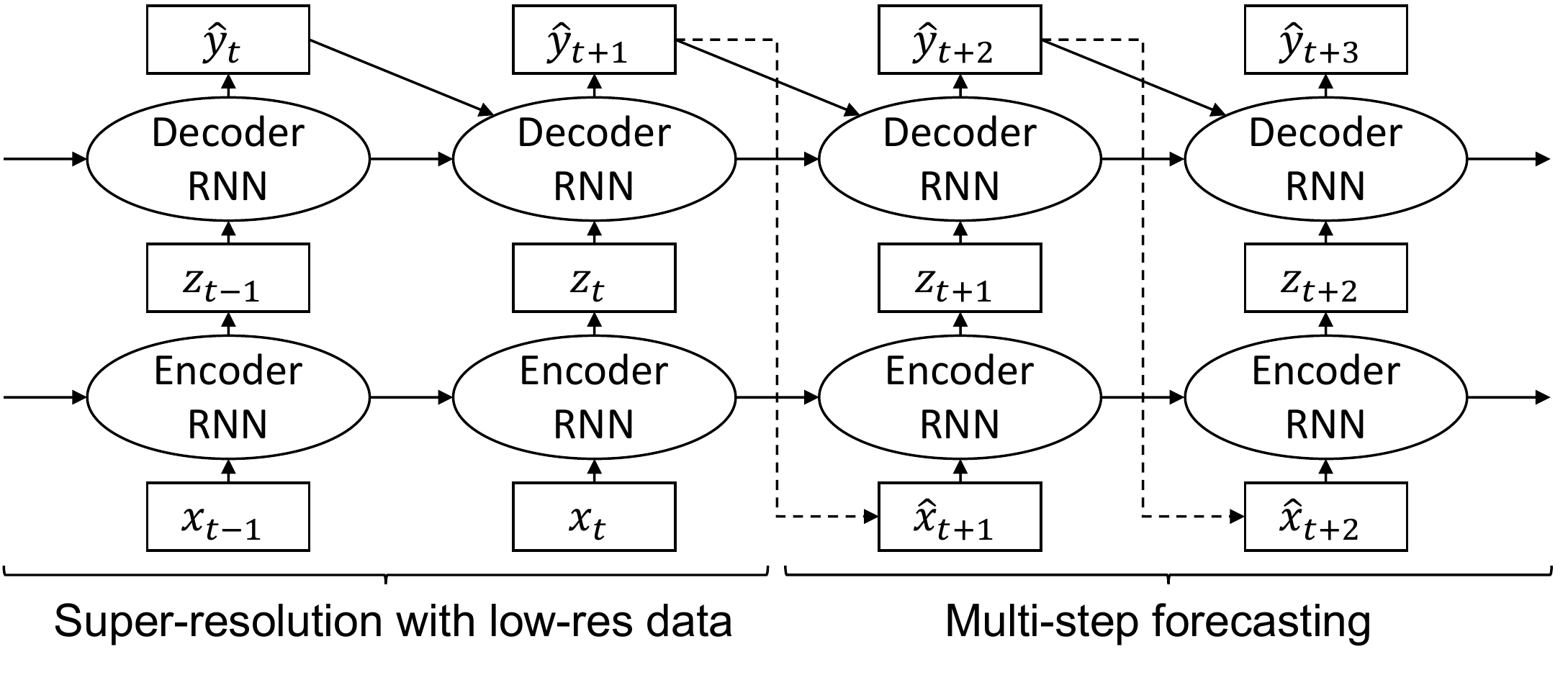}
    \caption{
        Full extrapolation model. The recurrent decoder takes as input the history of latent variables $z_{0:t}$ and the last HR output $\hat{y}_t$, and generates the new output $\hat{y}_{t+1}$.
    }
    \label{fig:network_extrap}
\end{figure}

As illustrated in Section~\ref{sec:arch} and Fig.~\ref{fig:network}, our model is set up to generate the super-resolution without prediction by default. However, with slight change of the mapping from $\bm x$ to $\bm z$ to $\bm y$, The model can be rerouted for for 1-step prediction. Specifically, instead of the following setup,
the following data generating distribution is defined,
\begin{equation}
    p(\bm{x}_{1:t+1}, \bm{z}_{0:t}) = \prod_{i=0}^t{p_\theta(\bm{x}_{i+1} | \bm{z}_i) p(\bm{z}_i | \bm{z}_{0:i-1})}.
    \label{eq:generative_extrap}
\end{equation}
We can then define a \textit{extrapolation} variant of the model. In other words, in the \textit{extrapolation model}, we aim to compute the probability distribution of the SR prediction $(c_{t+1},\bm{u}_{t+1})$, given the low resolution data at time $t$.  An illustration is shown in Fig.~\ref{fig:network_extrap}


\subsection{\textit{c}-only model}
In certain circumstances, we might want to do prediction of concentration given a reliable future wind forecasting. Such a configuration can be set up by repurposing the encoder to take as input $x'_{t} = (c_t, \bm{u}_{t+1})$, such that the data generating process can be written as, 
\begin{equation}
    p(\bm{x'}_{1:t}, \bm{z}_{0:t}) = \prod_{i=0}^t{p_\theta(\bm{x'}_{i} | \bm{z}_i) p(\bm{z}_i | \bm{z}_{0:i-1})}.
    \label{eq:generative_c_only}
\end{equation}
In this problem setup, we are interested in the joint probability distribution of the SR reconstruction of $\bm{u}_{t+1}$ given the low resolution velocity field at $t+1$ and the SR prediction of $c_{t+1}$ given the low resolution concentration at time $t$.

\section{Physics error}
\label{sec:phys_error}
To evaluate the model's ability to make physically consistent predictions, we calculate the physics errors based on Eqs.~(\ref{eq:ade}) and (\ref{eq:div_free}). Concretely, the advection-diffusion error $\epsilon_\textrm{adv-diff}$ and the divergence-free error $\epsilon_\textrm{div}$ is defined as follows,
\begin{align}
    \epsilon_\textrm{adv-diff} &= \| \partial_t c + \nabla \cdot c \bm{u} - \nabla \cdot (\bm{K} \cdot \nabla c) - Q \|^2_2
    \label{eq:e_adv_diff}
    \\
    \epsilon_\textrm{div} &=  \| \nabla \cdot \bm{u} \|^2_2
    \label{eq:e_div}
\end{align}

For the S3RP model and the bicubic baseline, the physics errors are visualized in Fig.~\ref{fig:phys_error}. The 1st column is used calculate $\epsilon_\textrm{adv-diff}$ as described in Eq.~(\ref{eq:e_adv_diff}), and the 2nd column is used to calculate $\epsilon_\textrm{div}$ described in Eq.~(\ref{eq:e_div}). The model only slightly outperforms the baseline in $\epsilon_\textrm{adv-diff}$ due to the lack of knowledge of $\bm K$ and $Q$. Whereas it significantly outperforms the baseline in $\epsilon_\textrm{div}$.


\section{Uncertainty and error}
\label{sec:error_std}

Our model is a generative model of which output is a sample from the probability distribution. Unlike a deterministic model, it is not straightforward to assess the accuracy of the probabilistic model for a high-dimensional spatio-temporal process. We have provided a quantitative measure in Table~\ref{table:compare_interp} in terms of the empirical coverage probability (ECP). It is shown that, while ECP for $1\sigma$ has a small error (only about 2.5\%), for $2\sigma$, there is about 12\% error in the coverage. This may due to the fact that the probability distribution is inferred from the LR data, while the comparison is made against the HR ground truth.

\begin{figure}[htbp]
    \centering
    \includegraphics[width=0.6\textwidth]{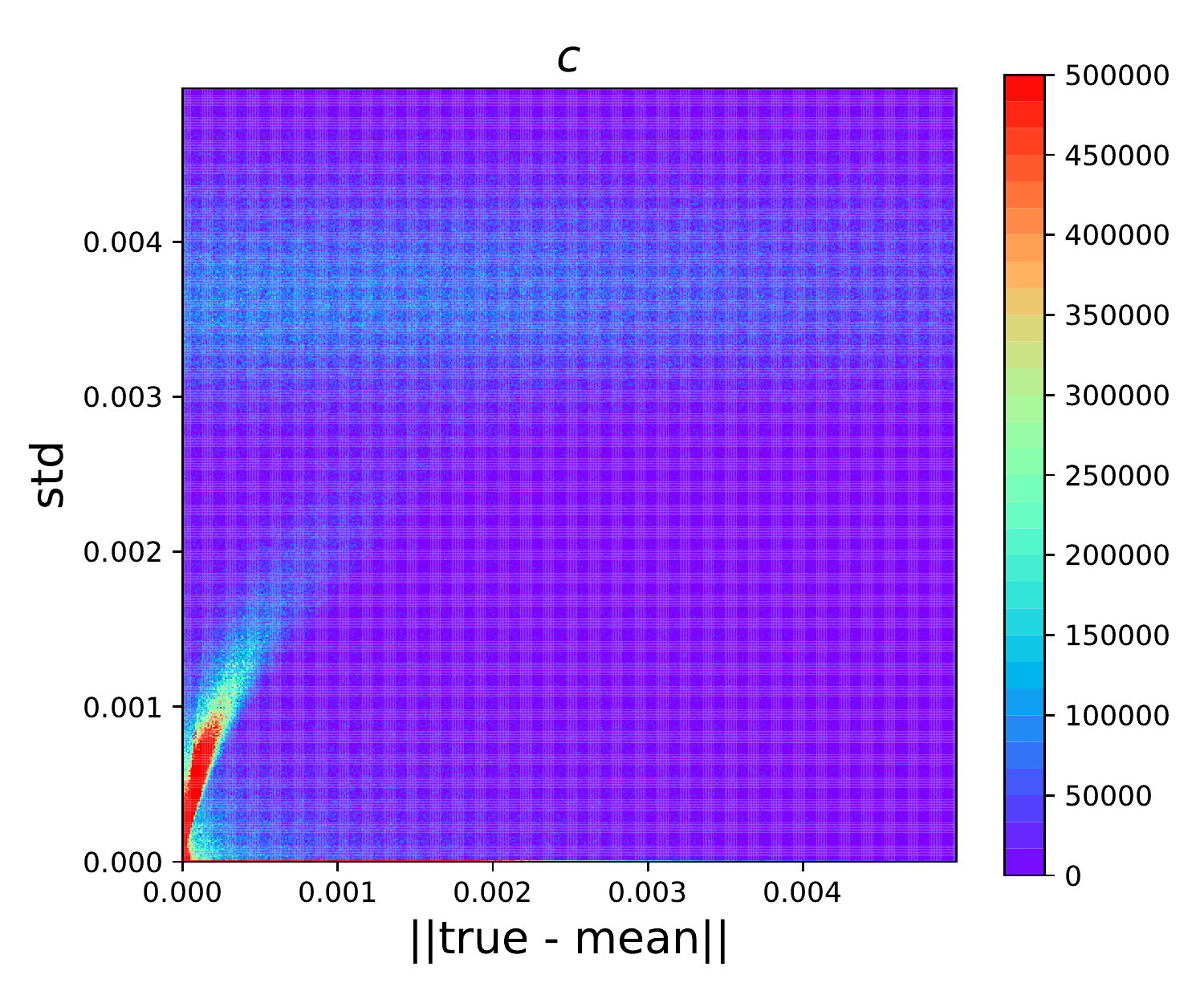}
    \caption{The standard deviation of model prediction plotted against the error in a 2D histogram. The colors correspond to the frequency density. An approximate trend that larger error corresponding to larger standard deviation indicates that the model performs well in confidence interval estimation.}
    \label{fig:error_std_appendix}
\end{figure}

Here, we provide another method to assess the estimated probability distribution from S3RP. Here, the probability distribution captures the uncertainty of the model prediction. In general, we expect that the width of the probability distribution increases when the model is uncertain about the state of the process. In other words, we expect to see a wider distribution, where the model error is larger. In Fig.~\ref{fig:error_std_appendix}, a two-dimensional histogram is plotted to compare a point-wise absolute difference, $| c_t(\bm{s}) - c^*_t(\bm{s})|$, in which $c_t(\bm{s})$ and $c^*_t(\bm{s})$ are, respectively, the model output and the HR ground truth at a spatial location ($\bm{s}$) at time $t$, and the width of the probability distribution, denoted by the standard deviation. It is shown that in general as $| c_t(\bm{s}) - c^*_t(\bm{s})|$ increases, the estimated probability distribution becomes wider. There is a faint band at around standard deviation between $0.003$ and $0.004$, which is not well understood. It may come from the error in the estimation of the probability distribution. Again, here we aim to estimate the probability distribution of the SR solution solely from the LR data, which makes the problem challenging.

\end{appendices}
\clearpage

\end{document}